\documentclass[10pt]{IEEEtran}
\IEEEoverridecommandlockouts
\usepackage{cite}
\usepackage{amsmath,amssymb,amsfonts}
\usepackage{graphicx}
\usepackage{float}
\usepackage{textcomp}
\usepackage{subcaption}
\usepackage{amsmath}
\usepackage{booktabs} 
\usepackage{multirow}
\usepackage[font=footnotesize,skip=0pt]{caption}

\usepackage{hyperref}       
\usepackage{url}            
\usepackage{caption} 
\usepackage{booktabs}       
\usepackage{amsfonts}       
\usepackage{nicefrac}       
\usepackage{microtype}      
\usepackage{xcolor}         
\usepackage{algorithm}
\usepackage{algpseudocode}
\usepackage{amsmath}
\usepackage{tikz}

\usepackage{orcidlink}
\usepackage{comment}
\usepackage{amssymb}
\usepackage{braket}
\usepackage{xcolor}
\usepackage{stfloats}
\usepackage{adjustbox}

\usepackage{setspace}
\usepackage{enumitem}
\setlist[itemize]{leftmargin=*}%
\setlist[enumerate]{leftmargin=*}%

\usepackage{algorithm}
\usepackage{algpseudocode}
\usepackage{amsmath} 
\usepackage{cuted}
\def\BibTeX{{\rm B\kern-.05em{\sc i\kern-.025em b}\kern-.08em
    T\kern-.1667em\lower.7ex\hbox{E}\kern-.125emX}}

\usepackage[compact]{titlesec}
\usepackage{stfloats}

\titlespacing\section{0pt}{0.2\baselineskip}{0.12\baselineskip}
\titlespacing\subsection{0pt}{0.15\baselineskip}{0.08\baselineskip}
\titlespacing\subsubsection{0pt}{0.1\baselineskip}{0.08\baselineskip}

\usepackage{fancyhdr}
\fancypagestyle{firstpage}{
   \fancyhf{} 
   \fancyhead[C]{To appear at 2026 IEEE International Joint Conference on Neural Networks (IJCNN) Proceedings, Maastricht, Netherlands} 
}

\begin{document}

\title{Design Space Exploration of Hybrid Quantum Neural Networks for Chronic Kidney Disease
}


\author{\IEEEauthorblockN{Muhammad Kashif \IEEEauthorrefmark{1}\IEEEauthorrefmark{2}, Hanzalah Mohamed Siraj\IEEEauthorrefmark{1}\IEEEauthorrefmark{2}, Nouhaila Innan\IEEEauthorrefmark{1}\IEEEauthorrefmark{2}, Alberto Marchisio\IEEEauthorrefmark{1}\IEEEauthorrefmark{2},
Muhammad Shafique\IEEEauthorrefmark{1}\IEEEauthorrefmark{2}}

\IEEEauthorblockA{\IEEEauthorrefmark{1} \normalsize eBrain Lab, Division of Engineering, New York University Abu Dhabi, PO Box 129188, Abu Dhabi, UAE\\}
\IEEEauthorblockA{\IEEEauthorrefmark{2} \normalsize Center for Quantum and Topological Systems, NYUAD Research
Institute, New York University Abu Dhabi, UAE}

Emails: \{muhammadkashif, sm12152, nouhaila.innan, alberto.marchisio,  muhammad.shafique\}@nyu.edu

\vspace{-15pt}
}

\maketitle
\pagestyle{empty}
\thispagestyle{firstpage}
\begin{abstract}
Hybrid Quantum Neural Networks (HQNNs) have recently emerged as a promising paradigm for near-term quantum machine learning. However, their practical performance strongly depends on design choices such as classical-to-quantum data encoding, quantum circuit architecture, measurement strategy and shots. 
In this paper, we present a comprehensive design space exploration of HQNNs for Chronic Kidney Disease (CKD) diagnosis. Using a carefully curated and preprocessed clinical dataset, we benchmark 625 different HQNN models obtained by combining five encoding schemes, five entanglement architectures, five measurement strategies, and five different shot settings. To ensure fair and robust evaluation, all models are trained using 10-fold stratified cross-validation and assessed on a test set using a comprehensive set of metrics, including accuracy, area under the curve (AUC), F1-score, and a composite performance score. Our results reveal strong and non-trivial interactions between encoding choices and circuit architectures, showing that high performance does not necessarily require large parameter counts or complex circuits. In particular, we find that compact architectures combined with appropriate encodings (e.g., IQP with Ring entanglement) can achieve the best trade-off between accuracy, robustness, and efficiency. Beyond absolute performance analysis, we also provide actionable insights into how different design dimensions influence learning behavior in HQNNs. 
 
\end{abstract}

\section{Introduction}

Quantum machine learning (QML) has emerged as a promising paradigm for leveraging near-term noisy intermediate-scale quantum (NISQ) devices in data-driven applications \cite{Schuld:2015,kashif:2021_DSE, zaman2023survey,innan2024financial,innan2025qnn,dutta2025quiet}. 
In QML, hybrid quantum–classical neural networks (HQNNs) are among the most practical and widely studied, where variational quantum circuits (VQCs) are embedded as trainable layers within classical neural network pipelines \cite{kashif:PP,innan_next_gen,kashif2025deep}. 
Such hybrid models aim to combine the representational power of quantum circuits with the optimization stability and scalability of classical deep learning \cite{kashif_demonstrating,kashif_adv}.

HQNNs also suffer from the infamous barren plateaus problem~\cite{mcclean2018barren}, where gradients vanish exponentially with size of underlying quantum circuits, however a number of strategies exist to mitigate or overcome these issues~\cite{kashif2024resqnets,kashif2024alleviating}. Similarly, the quantum hardware noise can also significantly deteriorate the performance of HQNNs. To this end, the performance robustness of HQNNs has also been widely investigated under realistic noise scenarios, as well as ways to exploit noise such that it enhances the training of HQNNs~\cite{kashif2024nrqnn,ahmed2025comparative,kashif2025hqnet, ahmed2025noisy}
Despite the rapid growth of HQNN development, their design is still largely based on heuristics. Existing studies typically fix one or two components, such as the encoding strategy or the VQC structure and report results on small benchmarks \cite{zaman2024studying,kashif2026faqnas}. However, an HQNN consists of multiple design stages that can significantly affect the overall performance of the model, including classical to quantum data encoding,  structure of VQC, type of qubit measurements, and the number of shots \cite{kashif:2021_DSE,innan2024financial1,el2026comparative}. These choices define a high-dimensional design space that strongly influences both performance and trainability, especially under NISQ constraints \cite{schuld_2021_effect,cerezo_exp,kashif_unified, marchisio2025cutting}. 

However, systematic design space exploration (DSE) of HQNNs for a given task remains largely unexplored.
%

In this paper, we consider Chronic Kidney Disease (CKD) classification, a clinically relevant medical diagnosis task involving heterogeneous, partially missing, and imbalanced data. Such datasets pose challenges not only for quantum models, but also for classical networks, making them an ideal test case for realistic hybrid quantum-classical pipelines.
Specifically, we conduct a systematic design space exploration framework for HQNNs for CKD classification by evaluating a large family of HQNN architectures instead of focusing on a single hand-crafted model.

\begin{figure}[h]
    \centering
    \includegraphics[width=0.85\linewidth]{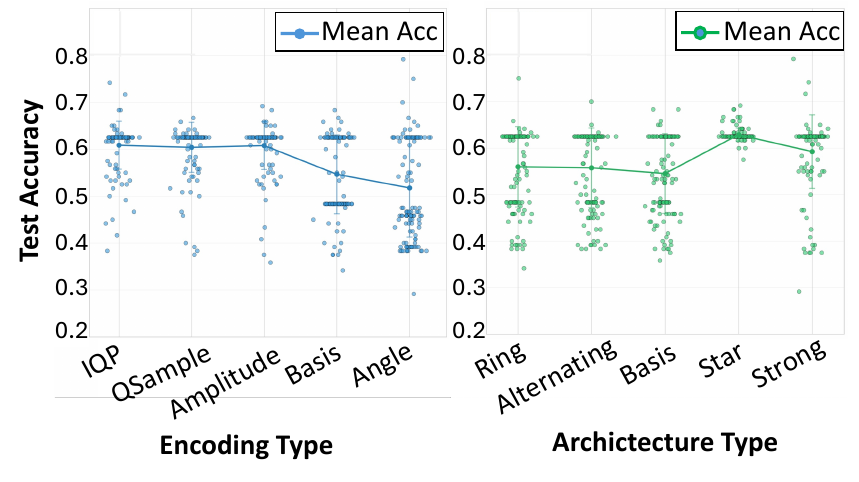}
    \caption{Motivational Analysis highlighting the need for Design Space Exploration of HQNNs. Only by changing encoding type or quantum circuit architecture (mainly differs in entanglement topology), the accuracy can vary between 30\% to 80\%.}
    \label{fig:mot_analysis}
\end{figure}
\subsection{Motivational Analysis}
Fig.~\ref{fig:mot_analysis} illustrates the classification accuracy obtained by HQNNs under different data encodings (left) and quantum circuit architectures (right), while keeping the dataset, training procedure, and optimization settings fixed. 
We observe high performance variability purely due to design choices. Depending on the selected encoding or quantum circuit design, the achieved accuracy can range from near-random performance (30–40\%) to relatively strong performance (75–80\%).

More importantly, these variations are not caused by changes in data, training budget, or optimizer, but purely by the architectural and encoding choices. Even among reasonable and commonly used encodings (e.g., IQP, amplitude, angle, basis) and quantum circuit architectures (e.g., ring, alternating, star, strong entangling), the spread in performance remains substantial. This demonstrates that HQNN performance is extremely sensitive to design decisions and that any arbitrary choices can easily lead to significantly suboptimal models.

These results highlight a critical challenge in HQNNs, i.e., there is no single universally good encoding or ansatz, and the performance landscape is highly non-trivial. Consequently, manual trial-and-error or heuristic architecture selection is unreliable and does not scale. This observation strongly motivates the need for systematic design space exploration for HQNNs, where quantum encoding, VQC structure, and model configurations must be treated as the primary optimization variables rather than fixed design choices.

\subsection{Our Contributions}

\begin{itemize}
    \item \textbf{Systematic HQNN Design Space Exploration.} We present a large-scale structured exploration of the HQNN design space, covering variations in data encoding, quantum circuit architectures, measurement strategies, and shot budgets within a unified framework (\textbf{Section~\ref{sec:methodology}}).

    \item \textbf{Unified Benchmarking Pipeline.} We introduce a standardized evaluation protocol based on stratified cross-validation, consistent data preprocessing, and controlled optimization settings, which enables fair and statistically robust comparison across all hybrid model configurations.

    \item \textbf{Comprehensive Real-World Case Study on CKD.} We provide the first extensive benchmark of HQNNs on a clinically relevant Chronic Kidney Disease dataset, moving beyond synthetic or toy problems, and identify practical performance and stability characteristics.
    
    \item \textbf{Multi-Metric Performance Analysis.} We evaluate models using a composite of clinically meaningful metrics and analyze the results to identify trade-offs between performance, robustness, and architectural complexity (\textbf{Section~\ref{sec:results}}).

    \item \textbf{Empirically Derived Design Guidelines for HQNNs.} Based on global trend analysis across the design space, we extract actionable insights and practical recommendations for designing HQNNs under NISQ constraints (\textbf{Section~\ref{sec:conclusion}}).
\end{itemize}


\section{Background}
\subsection{t-distributed Stochastic Neighbor Embedding (t-SNE)}
t-SNE is a non-linear dimensionality reduction technique for visualizing high-dimensional data in low-dimensional spaces \cite{anvari2025tsne}. It preserves local neighborhood structures by modeling pairwise similarities and uses a Student t-distribution to create a heavy-tailed distribution in the low-dimensional space.
We use t-SNE as a validation tool to ensure multi-dataset comparability. 

\subsection{Hybrid Quantum Neural Networks (HQNNs)}
HQNNs combine classical and quantum components, making them suitable for NISQ devices\cite{kashif2023impact}. The architecture consists of: (1) classical preprocessing, (2) classical-to-quantum data encoding layer, (3) VQC, (4) measurement layer, and (5) classical post-processing, as shown in Fig.~\ref{fig:HQNN_architecture}.
Below, we discuss all the primary quantum components of HQNNs:

\begin{figure}[h]
    \centering
    \includegraphics[width=1.0\linewidth]{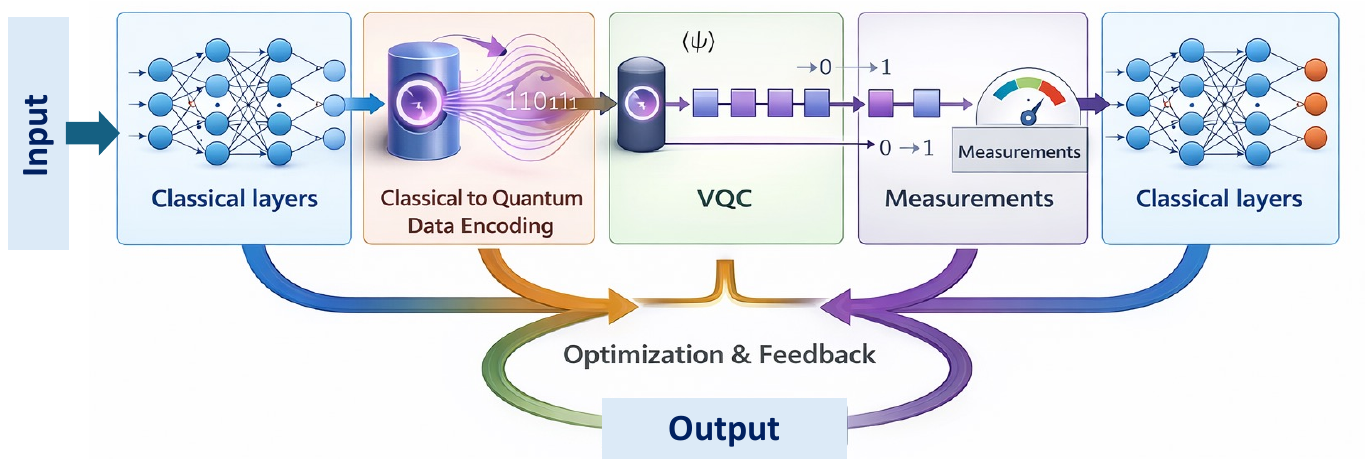}
    \caption{HQNN design flow, from input to output.}
    \label{fig:HQNN_architecture}
\end{figure}

\noindent
\textbf{Classical-to-Quantum Data Encoding Techniques:} \label{subsec:encodings}
There are various encoding methods, which differ in different qubit efficiency, expressivity, and hardware requirements. We use five encoding techniques:
\begin{enumerate}
    \item \textbf{Amplitude:} Maps features to quantum state amplitudes (exponential capacity: $2^n$ features with $n$ qubits)\cite{amp_enc}.
    \item \textbf{Angle:} Uses rotation gates to encode features as angles (hardware-friendly, one qubit per feature) \cite{ang_enc}.
    \item \textbf{Basis:} Maps discrete values to computational basis states (natural for binary/categorical features)\cite{basis_enc}.
    \item \textbf{IQP} encodes `n' features in `n' qubits using diagonal gates of an IQP circuit\cite{iqp_enc}. An IQP circuit is a quantum circuit of a block of Hadamards, followed by a block of gates that are diagonal in the computational basis.
    \item \textbf{QSample} combines amplitude and basis encoding~\cite{QSample1}.
    
\end{enumerate}
The choice of encoding can significantly affect the overall performance of HQNNs and is hence an important design choice in HQNNs.

\begin{figure*}
    \centering
    \includegraphics[width=0.9\linewidth]{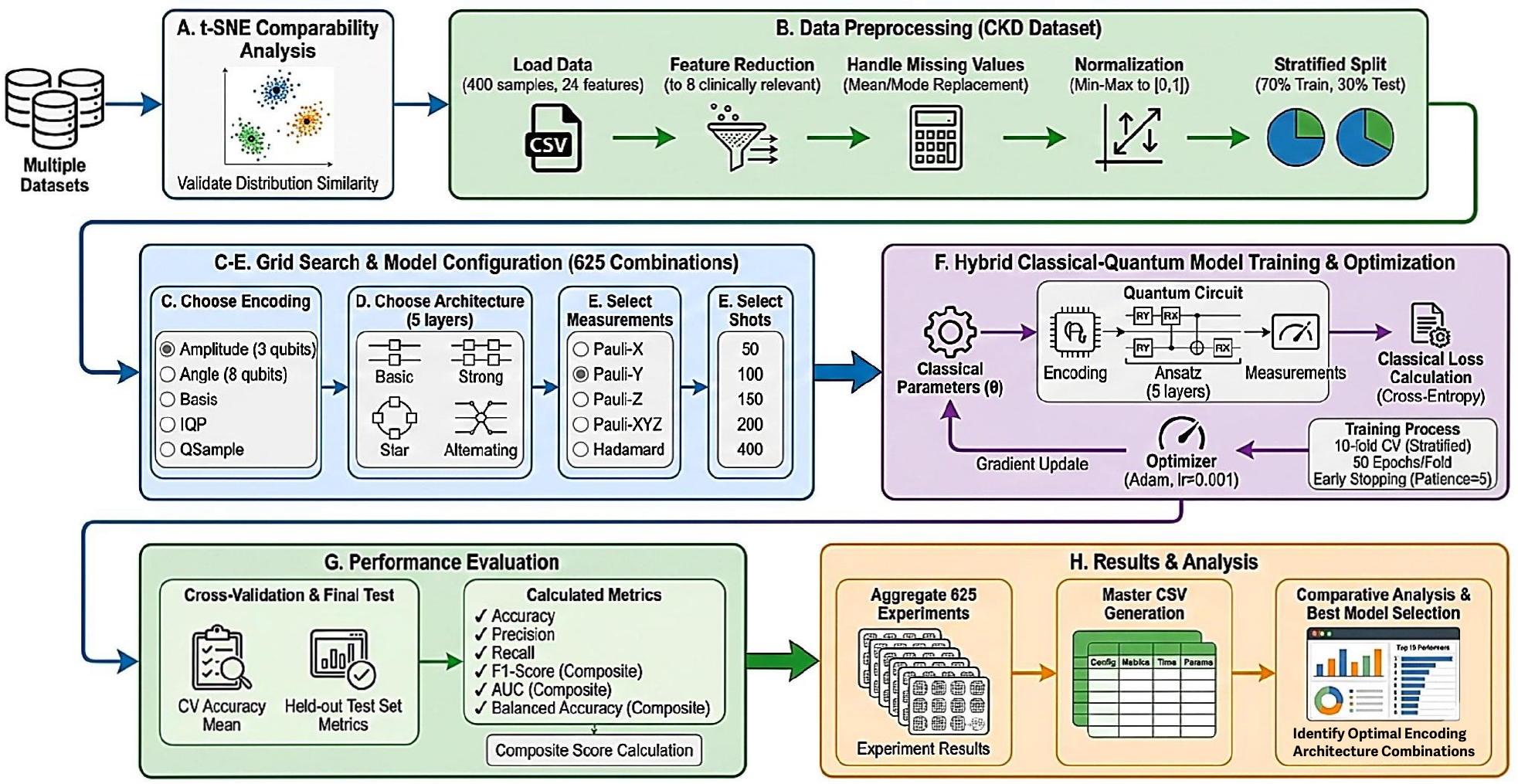}
    \caption{\footnotesize An overview of our methodology which includes dataset comparability analysis, data preprocessing, and a grid search over hybrid model configurations with varying encodings, quantum circuit architectures, measurements, and shot counts. Models are trained using a hybrid optimization loop with cross-validation and evaluated using composite performance metrics. Results are aggregated to identify the best-performing architectural configurations.}
    \label{fig:methodology}
\end{figure*}

\noindent
\textbf{Shots:} 
Expectation values of observables cannot be obtained from a single quantum circuit execution, but are estimated by repeatedly executing the same quantum circuit, a process known as \textit{sampling with shots}~\cite{phalak2023shot}. Each shot corresponds to one run of the circuit followed by a measurement, producing a single bitstring outcome. The final expectation value is then computed as the empirical average over all collected measurement outcomes. A larger number of shots reduces the statistical (sampling) noise and yields more accurate estimates, but increases the overall computational cost\cite{seksaria2025shots}. In hybrid quantum–classical algorithms, the number of shots therefore introduces a trade-off between measurement precision and training efficiency. Hence, shots are an important practical resource constraint on NISQ devices and an important design component in HQNNs.

\noindent
\textbf{Measurement Types:}
In HQNNs, qubit measurement serves as the interface between the quantum and classical components, as all information from the quantum circuit is extracted via expectation values of selected observables \cite{huggins2022nearly}. While measurements in the standard computational basis capture population statistics, they do not fully reveal phase and coherence information. Alternatively, measuring in multiple bases (e.g., Pauli-X, Pauli-Y, and Hadamard) provides complementary projections of the quantum state, increasing the expressive power of the model and revealing information that may be hidden in the case of single-basis measurements. Therefore, measurement choice is an important design dimension in HQNNs.


\section{Methodology}
\label{sec:methodology}

We perform a systematic design space exploration of HQNNs for CKD diagnosis, where performance is sensitive to specific quantum design choices. An overview of our methodology is presented in Fig.~\ref{fig:methodology}. We consider different CKD datasets, and evaluated 625 unique HQNN configurations, through a comprehensive grid search process. Each HQNN is trained under the same experimental settings and evaluated on multiple metrics, such as accuracy, AUC, F1-Score, and other composite scores made up of true and false positives/negatives, to identify the best configurations of HQNNs.

\begin{table}
\centering
\footnotesize
\caption{CKD datasets used in this study. DS=Dataset, Inst=Instances, Feat=Features, Miss=Missing values}
\label{tab:datasets}
\footnotesize
\begin{tabular}{lccccc}
\hline
\textbf{DS} & \textbf{Ref} & \textbf{Inst.} & \textbf{Feat.} & \textbf{Miss.} & \textbf{CKD / Not CKD (\%)} \\
\hline
D1 & \cite{Dataset-1/4} & 400  & 24 & Yes (Imputed) & 62.5 / 37.5 \\
D2 & \cite{Dataset-2}   & 1659 & 54 & No         & 91.9 / 8.9  \\
D3 & \cite{Dataset-3}   & 200  & 28 & No         & 64.0 / 36.0 \\
D4 & \cite{Dataset-1/4} & 400  & 24 & Yes (left empty)  & 62.5 / 37.5 \\
\hline
\end{tabular}
\end{table}

\subsection{Datasets}


We consider four CKD datasets summarized in Table~\ref{tab:datasets}, which differ in size, feature dimensionality, class balance, and data completeness. Dataset~2 is the largest and most imbalanced, whereas Datasets~1, 3, and 4 exhibit more moderate class distributions. Datasets~1 and~4 represent two variants of the same dataset, where Dataset~1 uses statistical imputation, i.e., mean for missing features because it preserves the overall statistical distribution of clinical features, which is important for medical data, and median for binary features (0/1). Dataset~4 leaves missing entries unfilled.
To ensure fair benchmarking across datasets, we assess dataset comparability using t-SNE analysis, as discussed in the following section.

\subsection{t-SNE-Based Dataset Comparability Analysis}

When multiple datasets are used, differences in data distribution and class imbalance can result in misleading performance comparisons. Therefore, to ensure that the observed variations in performance arise from model design choices rather than dataset bias, we assess dataset comparability using a t-SNE–based analysis with number of components set to 2, random seed fixed to 42, perplexity chosen as $\min(30, \text{samples}/4 )$, and the number of iterations set to 500. 
Dataset similarity is quantified via centroid distances of each dataset in the embedded space, which provides a proxy for global distributional alignment among the datasets. Based on this analysis, Datasets~1,~3, and~4 are found to be mutually comparable, with Datasets~1 and~4 being most similar, while Dataset~2 exhibits significant distributional discrepancy. Dataset~2 was initially included for completeness but excluded due to distributional mismatch.



\subsection{Data Preprocessing}


%
\subsubsection{Feature Reduction}
Due to the high number of features in the original datasets (Table ~\ref{tab:datasets}), and the limited scalability of quantum circuits on NISQ devices, we apply Principal Component Analysis (PCA) to reduce all datasets to 8 features, which results in efficient quantum resource utilization by using only 3 qubits ($2^3=8$) for amplitude encoding and 8 qubits for other encoding schemes with one qubit per feature. Additionally, feature reduction significantly improves the computational feasibility given the large search space of HQNN configurations in this paper.

\subsubsection{Feature Normalization}

The clinical features in the datasets we have used have vastly different scales (e.g., age ranges 2--90, blood pressure ranges 50--180, while specific gravity is 1.005--1.025). Without normalization, features with larger numerical ranges would dominate the quantum encoding, causing the quantum circuit to focus on scale rather than meaningful patterns.
We applied \texttt{Min-Max scaling} ($X_{scaled} = (X - X_{min}) / (X_{max} - X_{min})$) to transform all 8 features to the [0,1] range, ensuring equal representation in our quantum circuits.

\subsubsection{Data Splitting}
The datasets are split into 70\% training and 30\% testing. The test set remains completely unseen during all training and cross-validation, providing an unbiased estimate of how each HQNN model would perform on new CKD patients. We used scikit-learn's \texttt{train\_test\_split} with \texttt{test\_size=0.3} and \texttt{random\_state=42} to ensure reproducibility across all 625 experiments.



\subsubsection{StratifiedKFold sampling - Handling Class Imbalance} \label{subsec:stratified_sampling}

The datasets we used in this paper exhibit significant class imbalance (Table~\ref{tab:datasets}), which can result in biased performance estimates if not handled carefully. To address class imbalance, we use \texttt{StratifiedKFold} with \texttt{n\_splits=10} and \texttt{random\_state=42}, to preserve class proportions across all CV folds. Each of the 10 folds maintains the 62.5\%/37.5\% class distribution. This process ensures that HQNNs learn to recognize both CKD and healthy patients equally well, resulting in reliable performance estimates, across the 625 HQNN configurations.



\subsection{Grid Search and Model Configurations} 
To systematically identify optimal HQNN configurations for medical classification, we use an exhaustive grid search approach that evaluates all combinations of key design parameters. An overview of our grid search approach is presented in Algorithm ~\ref{algo_grid_search}. 

\begin{algorithm}
\footnotesize
\caption{Exhaustive Grid Search}
\label{algo_grid_search}
\begin{algorithmic}[1] 

    \For{\textit{each encoding in [Amplitude, Angle, Basis, IQP, QSample]}}
        \For{\textit{each architecture in [Basic, Ring, Star, Strong, Alternating]}}
            \For{\textit{each measurement in [Pauli-X, Y, Z, XYZ, Hadamard]}}
                \For{\textit{each shots in [50, 100, 150, 200, 400]}}
                    
                    \State \textit{Train HQNN model}
                     \State \textit{Adam optimizer (lr=0.001)}
                    
                    \State \textit{Evaluate and record performance metrics}
                    
                \EndFor
            \EndFor
        \EndFor
    \EndFor
    
    \Statex 
    \State Total experiments: $5 \times 5 \times 5 \times 5 = 625$ configurations
\end{algorithmic}
\end{algorithm}

\begin{figure}
    \centering
    \includegraphics[width=1.0\linewidth]{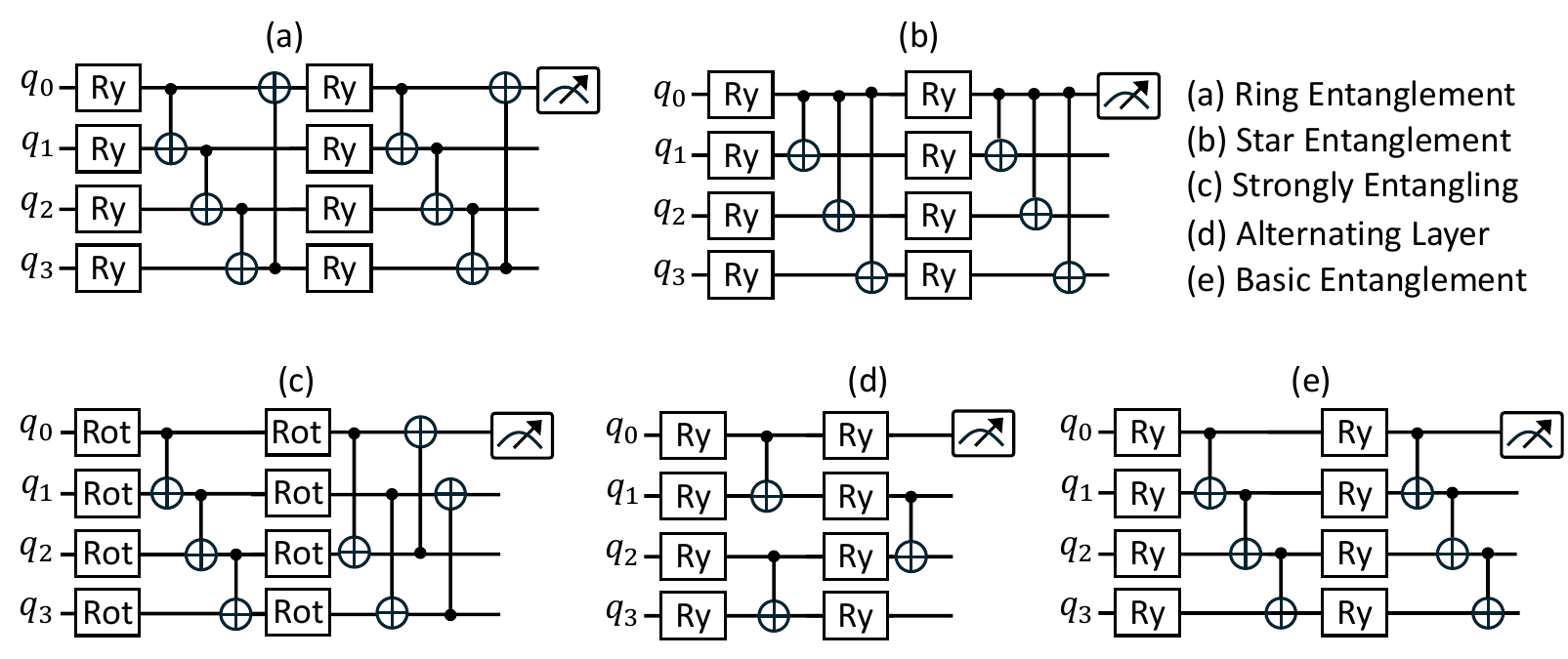}
    \caption{Quantum Circuit Architectures with different entanglement structures used.}
    \label{fig:ent_types}
\end{figure}

We perform an exhaustive grid search over 625 HQNN configurations by varying five data encodings (Amplitude, Angle, Basis, IQP, and QSample), five entanglement topologies (Basic, Ring, Strong, Alternating, and Star, as shown in Fig.~\ref{fig:ent_types}), five measurement bases (Pauli-X, Pauli-Y, Pauli-Z, Pauli-XYZ, and Hadamard), and five shot settings (50, 100, $\cdots$400). 
%
To ensure fair comparison across all HQNN configurations, the quantum circuit depth is fixed to 5 layers, each consisting of parameterized single-qubit rotations followed by architecture-specific entangling gates, resulting in approximately 15–40 trainable parameters depending on the topology.

\subsection{Hybrid Classical-Quantum Model Training \& Optimization}

HQNNs are trained using a resource-aware design in which the number of qubits is adapted to the data encoding strategy. Amplitude encoding uses three qubits to represent eight features, while Angle, Basis, IQP, and QSample encodings use eight qubits with one feature per qubit, ensuring efficient and consistent feature representation across models.

For a fair comparison, all HQNN configurations in the search space are trained under a unified protocol based on cross-validation, early stopping, and final testing on a held-out dataset. 
We employ 10-fold stratified cross-validation with early stopping and evaluate performance using the mean cross-validation accuracy as part of the composite score. This cross-validation accuracy serves as one component of our composite performance score (more details in the following Section~\ref{subsec:performance_eval}), and helps in identifying quantum circuit architectures that generalize consistently rather than performing well only on a single favorable split.

Training is performed for 50 epochs using mini-batches of size 16. Adam optimizer with a learning rate of 0.001 is used for optimization. Moreover, early stopping with a patience of 5 epochs based on the validation loss is used to prevent overfitting. Gradients of quantum circuits are computed using PennyLane’s automatic differentiation framework based on parameter-shift rules~\cite{bergholm:2022pennylane}. At each training step, the gradients of the loss function with respect to the circuit parameters are evaluated and used to update the model parameters. This process is repeated for all batches and epochs until convergence or early stopping is triggered.

\subsection{Performance Evaluation} \label{subsec:performance_eval}

We evaluate HQNNs performance using single and composite metrics. For single metrics we use accuracy, precision, recall, and F1-score (Eqs.~\ref{eq:acc}-\ref{eq:f1_score}). While accuracy reflects the overall correctness of predictions (avoiding false positives, i.e., incorrectly diagnosing healthy patients) and recall (avoiding false negatives, i.e., missing actual CKD cases), they are particularly important for medical diagnosis due to the asymmetric costs of false positives and false negatives, with the F1-score providing a balanced summary of both precision and recall.

{\footnotesize
\begin{equation}\label{eq:acc}
Accuracy = \frac{TP + TN}{TP + TN + FP + FN}
\end{equation}

\begin{equation}\label{eq:precision}
Precision = \frac{TP}{TP + FP}
\end{equation}

\begin{equation}\label{eq:recall}
Recall = \frac{TP}{TP + FN}
\end{equation}

\begin{equation}\label{eq:f1_score}
F1 = \frac{2 (Precision \cdot Recall)}{Precision + Recall}
\end{equation}
}

\subsubsection{Composite Performance Metrics} 

While individual metrics (accuracy, precision, recall, F1) provide valuable insights, they each capture only one aspect of model performance. For our comprehensive evaluation of 625 HQNN configurations, it is essential to have composite metrics that combine multiple performance dimensions to identify the best overall configurations.
Therefore, we used multiple composite metrics, which we discuss below:

\paragraph{\textbf{MCC-F1 Curves}} Matthews Correlation Coefficient (MCC) is a correlation coefficient between observed and predicted classifications, ranging from -1 to +1, where +1 indicates perfect prediction, 0 indicates random prediction, and -1 indicates perfect inverse prediction~\cite{mcc-f1score}. MCC (Eq.~\ref{eq:MCC}) combined with F1-score provides a balanced assessment that accounts for all four confusion matrix elements (TP, TN, FP, FN), making it particularly suitable for imbalanced datasets. MCC-F1 curves evaluate performance across different thresholds, revealing how the trade-off between MCC and F1 varies.

{\footnotesize
\begin{equation}\label{eq:MCC}
\mathrm{MCC} =
\frac{TP \cdot TN - FP \cdot FN}
{\sqrt{(TP + FP)\,(TP + FN)\,(TN + FP)\,(TN + FN)}}
\end{equation}
}

\paragraph{\textbf{General Performance Score (GPS)}} In order to comprehensively evaluate the HQNNs performance, we further use four GPS variants (GPS1-GPS4), proposed in ~\cite{gps}. These metrics combine different sets of metrics using harmonic means. GPS provides a single composite score that balances multiple performance aspects, enabling direct ranking of all 625 configurations. Different GPS formulations allow us to emphasize different metric combinations (e.g., GPS1 focuses on balanced accuracy, Area under the ROC curve (AUC), and F1, while GPS4 includes cross-validation accuracy). Using AUC and balanced accuracy makes composite metrics even more reliable. This is AUC because it measures how well the model distinguishes CKD from healthy patients regardless of threshold choice. Similarly, balanced accuracy defined as, $(TPR + TNR) / 2$ where $TPR = TP/(TP+FN)$ and $TNR = TN/(TN+FP)$, is crucial in imbalanced datasets, as in our case of CKD dataset, because it accounts for the class imbalance by giving equal weight to correctly identifying both healthy and diseased patient.   
All GPS metrics are presented in Eq.~\ref{eq:gps1}-\ref{eq:gps4}, more details on these composite metrics in ~\cite{gps}.

\begin{equation}\label{eq:gps1}
    \text{GPS1} = \frac{3}{\frac{1}{\text{test balanced accuracy}} + \frac{1}{\text{test auc}} + \frac{1}{\text{test f1}}}
\end{equation}

\begin{equation}\label{eq:gps2}
    \text{GPS2} = \frac{4}{\frac{1}{\text{test accuracy}} + \frac{1}{\text{test precision}} + \frac{1}{\text{test recall}} + \frac{1}{\text{test f1}}}
\end{equation}

\begin{equation}\label{eq:gps3}
    \text{GPS3} = \frac{3}{\frac{1}{\text{test balanced accuracy}} + \frac{1}{\text{test auc}} + \frac{1}{\text{test accuracy mean}}}
\end{equation}

\begin{equation}\label{eq:gps4}
    \text{GPS4} = \frac{4}{\frac{1}{\text{test balanced accuracy}} + \frac{1}{\text{test auc}} + \frac{1}{\text{test f1}} + \frac{1}{\text{cv accuracy mean}}}
\end{equation}

\paragraph{Sensitivity-Specificity Curves} In addition to individual and composite metrics discussed above, we also analyze the sensitivity (Eq.~\ref{eq:recall}) and specificity (Eq.~\ref{eq:specificity}) curves to evaluate the trade-off between sensitivity and specificity across different thresholds, which is critical for medical diagnosis where both correctly identifying CKD patients and avoiding false alarms for healthy patients are important~\cite{sensitivity}.

{\footnotesize
\begin{equation} \label{eq:specificity}
    \text{Specificity} = \frac{TN}{TN + FP}
\end{equation}
}

\section{Results and Discussion}\label{sec:results}
For each dataset, we summarize the 625-run design space using the same set of views to avoid repeating factor definitions. We first report factor-averaged GPS trends to highlight which encodings and entanglement topologies (architectures) are preferred under composite objectives. We then examine the accuracy distribution across factors (encoding, architecture, measurement, and shot budget) to assess robustness and dispersion beyond the single best run. Next, we relate MCC-F1 to capture the balance between precision/recall and correlation-based agreement, and visualize the specificity-sensitivity trade-off. Finally, we list the strongest configurations and provide an overlap table that aggregates the Top-5 lists across metrics, retaining only configurations that recur at least three times (Count $\ge 3$) to emphasize consistently competitive design choices.
\subsection{Dataset 1 Analysis}
\subsubsection{Factor-averaged GPS trends (encoding and architecture)}
\begin{figure*}[htpb]
    \centering
    \includegraphics[width=0.95\linewidth]{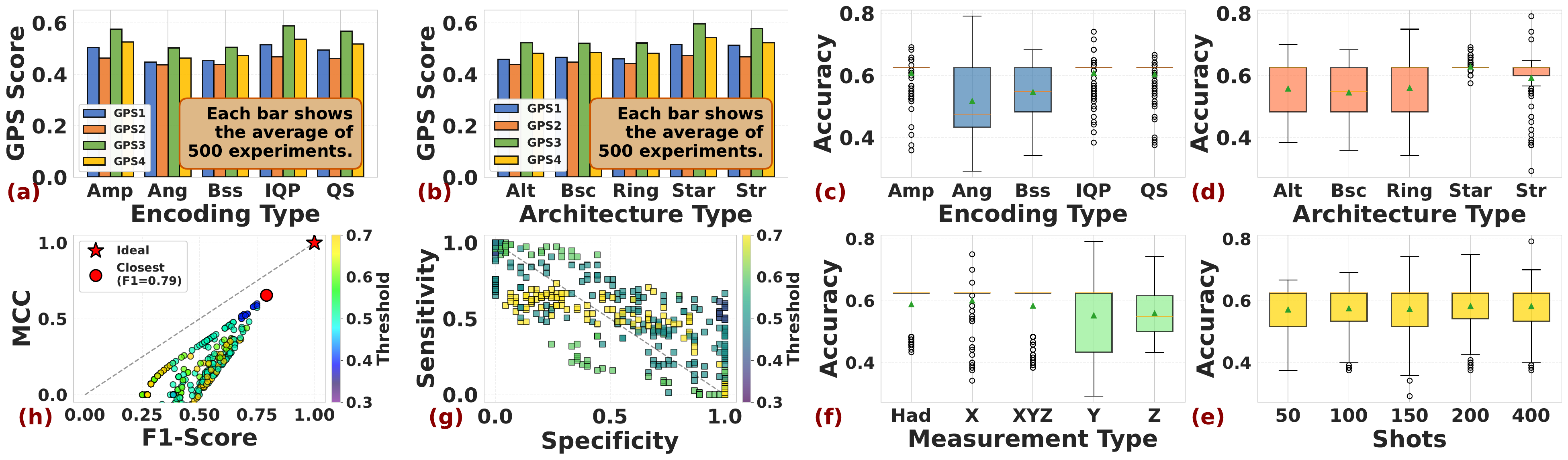}
    \caption{\textbf{Dataset 1:} Summary of the 625-configuration HQNN design grid. (a) Factor-averaged GPS by encoding. (b) Factor-averaged GPS by entanglement topology (architecture). (c) Accuracy distribution by encoding. (d) Accuracy distribution by architecture. (e) Accuracy distribution by shot budget. (f) Accuracy distribution by measurement observable. (g) Specificity vs sensitivity across configurations. (h) MCC vs F1 across configurations.}
    \label{dataset1}
\end{figure*}

As shown in Fig.~\ref{dataset1}-a, IQP attains the highest factor-averaged GPS scores across GPS1--GPS4 (often tied for the top rank), with Amplitude and QSample forming a second tier, while Angle and Basis remain lower on average. This pattern suggests that encodings inducing richer quantum feature maps preserve more task-relevant structure than direct per-feature rotation schemes in this design space. Fig.~\ref{dataset1}-b shows a consistent architecture effect: Star achieves the highest mean GPS values (most clearly for GPS3 and GPS4), followed by Strong, whereas Alternating, Basic, and Ring cluster lower and remain relatively close. Overall, the factor-averaged trends indicate that stronger connectivity increases aggregate performance, while differences among the simpler topologies are comparatively small under these objectives.

\subsubsection{Accuracy distribution and robustness across factors}
Accuracy distributions show that stability varies by factor. By encoding, Angle exhibits the widest spread and lower median, whereas IQP and QSample shift the distribution upward with more runs concentrated near the top range; Amplitude is tighter but slightly below the top tier. By architecture, Star and Strong raise medians relative to Alternating/Basic/Ring but still include low outliers. Measurement choice has the effect on dispersion: Hadamard, Pauli-X, and Pauli-XYZ remain tightly concentrated, while Pauli-Y produces larger variability and a lower-centered distribution, with Pauli-Z intermediate. Shot counts (50-400) induce only modest median changes compared to the above factors. Results are shown in Fig. \ref{dataset1}-(c-f).
\subsubsection{MCC-F1 and specificity-sensitivity}
Higher F1 is generally accompanied by higher MCC, but intermediate F1 values map to a wide range of MCC scores, indicating that configurations with similar F1 can differ substantially in error symmetry (i.e., the balance of false positives vs false negatives). The specificity-sensitivity plane exhibits a clear tradeoff structure, and threshold variation mainly shifts operating points along this curve rather than yielding a uniformly balanced region across the grid (Fig.~\ref{dataset1}-(g--h)).

\subsubsection{Top configurations}
Table~\ref{tab:d1_top5_overlap_values_c3} shows that Angle/Strong/Pauli-Y/400 is the most consistently selected configuration, appearing in the Top-5 of all metrics (Count=7) and ranking first for all objectives except GPS3 where it remains Top-3, indicating strong agreement between GPS1--2 and GPS3--4. The two other recurrent settings (Count=6) reflect different alignments: Angle/Ring/Pauli-X/200 is competitive for accuracy and GPS1--2 but does not appear under GPS3 and is weaker for GPS4, whereas IQP/Strong/Pauli-Z/150 remains competitive for GPS1--2 and ranks second for both GPS3 and GPS4. Overall, the overlap pattern highlights Angle/Strong/Pauli-Y as the most stable choice, with IQP/Strong/Pauli-Z emerging as the primary alternative under balanced composites.

\begin{table*}[htpbt]
\centering
\footnotesize
\caption{\textbf{Dataset 1:} Overlap of Top-5 configurations across metrics (Count $\ge 3$), reported as rank(value); ``--'' indicates absence from that metric's Top-5. Rows are sorted by Count in descending order.}
\label{tab:d1_top5_overlap_values_c3}
\begin{adjustbox}{max width=\textwidth}
\begin{tabular}{l c c c c c c c c}
\hline
\textbf{Config (Enc / Arch / Meas / Shots)} & \textbf{Count} &
\textbf{Acc} & \textbf{MCC-F1} & \textbf{Sens-Spec} & \textbf{GPS1} & \textbf{GPS2} & \textbf{GPS3} & \textbf{GPS4} \\
\hline
Angle / Strong / Pauli-Y / 400 & 7 &
1(0.7917) & 1(0.7147) & 1(0.7643) & 1(0.8466) & 1(0.8090) & 3(0.7321) & 1(0.7460) \\
Angle / Ring / Pauli-X / 200 & 6 &
2(0.7500) & 5(0.6503) & 2(0.7341) & 3(0.7644) & 2(0.7679) & -- & 5(0.6922) \\
IQP / Strong / Pauli-Z / 150 & 6 &
3(0.7417) & 4(0.6570) & -- & 2(0.8059) & 3(0.7673) & 2(0.7369) & 2(0.7381) \\
IQP / Strong / Pauli-Z / 200 & 4 &
4(0.7167) & -- & 4(0.7197) & 4(0.7486) & 4(0.7281) & -- & -- \\
\hline
\end{tabular}
\end{adjustbox}
\end{table*}

\subsection{Dataset 3 Analysis}
\begin{figure*}[htpb]
    \centering
    \includegraphics[width=0.95\linewidth]{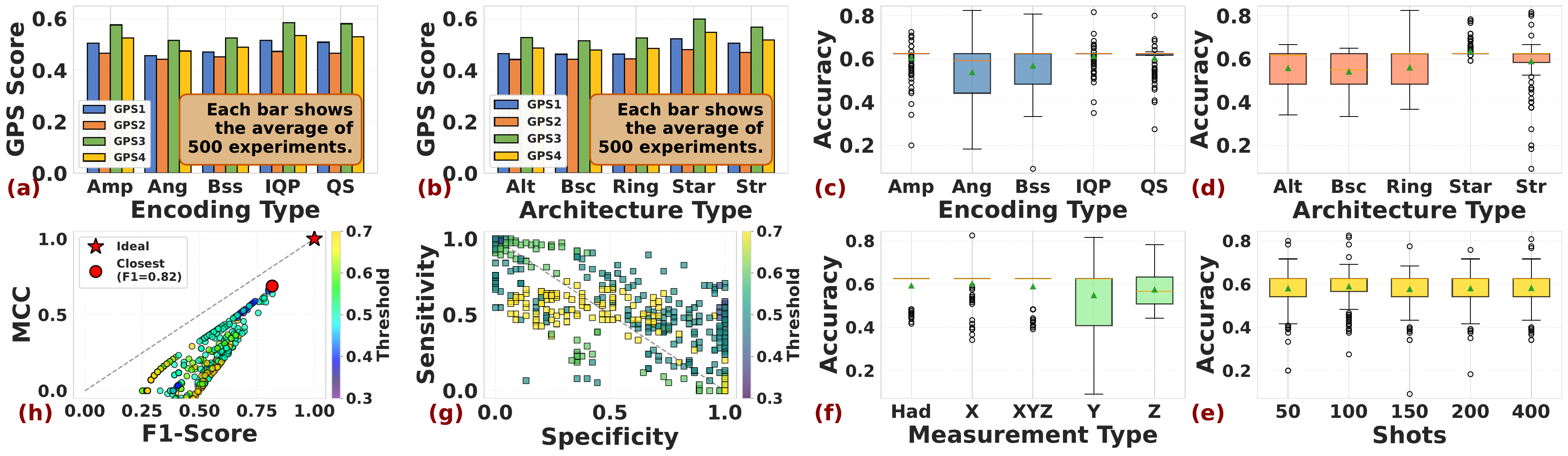}
    \caption{\textbf{Dataset 3:} Summary of the 625-configuration HQNN design grid (same panel order as Dataset 1).}
    \label{dataset3}
\end{figure*}
\subsubsection{Factor-averaged GPS trends (encoding and architecture)}
IQP and QSample provide the strongest GPS behavior overall, while Angle and Basis are typically lower on average; Amplitude remains competitive depending on the GPS variant. For entanglement, Star yields the highest mean GPS values, with Strong also performing well, whereas Alternating/Basic/Ring form a lower, tighter cluster. These trends are summarized in Fig. \ref{dataset3}-(a-b).
\subsubsection{Accuracy distribution and robustness across factors}
Fig. \ref{dataset3}-(c-f) indicates that variability is driven by model and measurement design than by shot budget. Encodings differ in median accuracy and in dispersion, with the best-performing encodings showing fewer low-end failures. Star and Strong architectures shift the distribution upward relative to simpler patterns, but outliers remain, indicating that high average performance does not remove unstable configurations. Measurement choice introduces the stability contrast, where some observables yield tight, high-accuracy distributions while others produce a wider spread and more low-accuracy runs. Changing shots from 50-400 produces modest shifts in the median relative to these effects.

\subsubsection{MCC-F1 and specificity-sensitivity}
The MCC vs F1 plot (Fig. \ref{dataset3}-g) represents a consistent upward trend, but with a broad middle region where similar F1 values map to noticeably different MCC values, reflecting confusion-matrix asymmetry across configurations. The specificity-sensitivity scatter (Fig. \ref{dataset4}-h) highlights a pronounced operating-point tradeoff across the grid, with threshold variation influencing where configurations lie on this curve rather than producing a single uniformly optimal region.
\subsubsection{Top configurations}
Table~\ref{tab:d3_top5_overlap_values_c3} shows that high-ranking configurations in Dataset~3 are distributed across several design points rather than concentrating on a single setting. Angle/Ring/Pauli-X/100 and IQP/Strong/Pauli-Y/100 each appear in five Top-5 lists and primarily support accuracy-, MCC-F1-, and GPS1--2-oriented objectives. In contrast, GPS3--4 are dominated by Basis/Star/Pauli-Z configurations (shots 100 and 150), indicating a shift toward a different region of the design space under the balanced composites. QSample/Strong/Pauli-Y/50 remains competitive for GPS1--2 and MCC-F1 (Count=4) but does not persist under GPS3--4, reinforcing the separation between these objectives.

\begin{table*}[htpbt]
\centering
\footnotesize
\caption{\textbf{Dataset 3:} Overlap of Top-5 configurations across metrics.}
\label{tab:d3_top5_overlap_values_c3}
\begin{adjustbox}{max width=\textwidth}
\begin{tabular}{l c c c c c c c c}
\hline
\textbf{Config (Enc / Arch / Meas / Shots)} & \textbf{Count} &
\textbf{Acc} & \textbf{MCC-F1} & \textbf{Sens-Spec} & \textbf{GPS1} & \textbf{GPS2} & \textbf{GPS3} & \textbf{GPS4} \\
\hline
Angle / Ring / Pauli-X / 100 & 5 &
1(0.8250) & 4(0.7127) & 1(0.8244) & 2(0.8526) & 2(0.8198) & -- & -- \\
IQP / Strong / Pauli-Y / 100 & 5 &
2(0.8167) & 1(0.7444) & 2(0.7926) & 1(0.8665) & 1(0.8301) & -- & -- \\
Basis / Star / Pauli-Z / 100 & 5 &
-- & 5(0.6901) & -- & 5(0.8063) & 4(0.7961) & 1(0.7905) & 1(0.7884) \\
QSample / Strong / Pauli-Y / 50 & 4 &
4(0.8000) & 2(0.7245) & -- & 3(0.8515) & 3(0.8160) & -- & -- \\
Angle / Star / Pauli-Z / 50 & 4 &
5(0.7833) & -- & 4(0.7781) & -- & -- & 4(0.7789) & 3(0.7796) \\
Basis / Strong / Pauli-Y / 400 & 3 &
3(0.8083) & -- & -- & 4(0.8133) & 5(0.7948) & -- & -- \\
Basis / Star / Pauli-Z / 150 & 3 &
-- & -- & 5(0.7742) & -- & -- & 2(0.7890) & 2(0.7850) \\
\hline
\end{tabular}
\end{adjustbox}
\end{table*}

\subsection{Dataset 4 Analysis}
\begin{figure*}[b]
    \centering
    \includegraphics[width=0.95\linewidth]{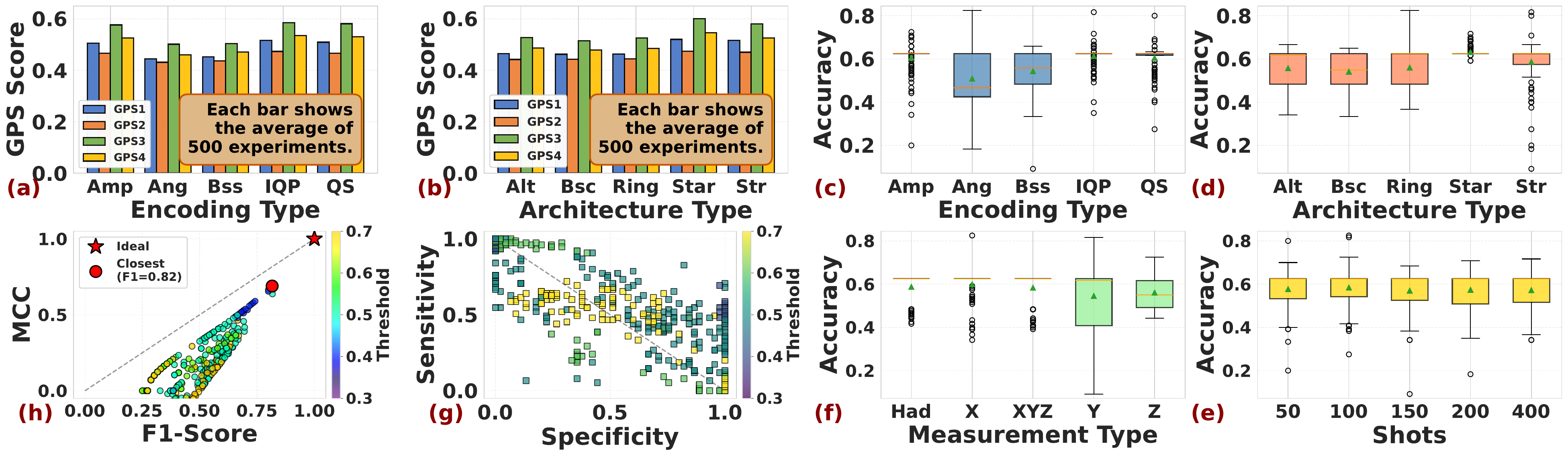}
    \caption{\textbf{Dataset 4:} Summary of the 625-configuration HQNN design grid (same panel order as Dataset 1).}

    \label{dataset4}
\end{figure*}
\subsubsection{Factor-averaged GPS trends (encoding and architecture)}
As presented in Fig.~\ref{dataset4}-(a--b), the composite objectives separate into two preference regimes. GPS1--2 favor IQP and QSample, whereas GPS3--4 increase the weight of Star-based settings, consistent with the configurations that repeatedly appear with Pauli-Z measurement in the Top-5 lists. Across architectures, Strong and Star achieve the highest mean GPS values, while Alternating/Basic/Ring remain lower and close to each other, indicating that stronger connectivity is the main differentiator at the topology level.

\subsubsection{Accuracy distribution and robustness across factors}
Fig. \ref{dataset4}-(c--d) shows that accuracy is primarily shaped by encoding and architecture choices, with Strong/Star shifting the distribution upward relative to Alternating/Basic/Ring. The encoding-wise box plot indicates that competitive performance is concentrated in a subset of encodings, whereas the rest exhibit lower medians and/or larger dispersion. Measurement choice further modulates robustness and reveals noticeably different spreads across observables (Fig. \ref{dataset4}-f), while the shot-wise box plot (Fig. \ref{dataset4}-e) suggests that increasing shots yields comparatively smaller distributional changes than selecting the encoding, architecture, and measurement.

\subsubsection{MCC-F1 and specificity-sensitivity}
The diagnostic panels in Fig.~\ref{dataset4}-(g--h) clarify how accuracy relates to error structure. MCC increases with F1 but exhibits a visible spread for intermediate F1, indicating that configurations with similar F1 can differ in their balance of false positives and false negatives. Similarly, the specificity-sensitivity plot shows a pronounced tradeoff curve, with threshold variation shifting operating points along the curve rather than collapsing them into a single balanced region. These patterns motivate reporting MCC and sensitivity-specificity alongside accuracy when selecting HQNN configurations.

\subsubsection{Top configurations}
Table~\ref{tab:d4_top5_overlap_values_c3} shows substantial cross-metric consistency, dominated by Strong connectivity with Pauli-Y measurement. QSample/Strong/Pauli-Y/50 is the only configuration appearing in all Top-5 lists (Count=7) and remains competitive under both point metrics and the GPS3--4 composites. IQP/Strong/Pauli-Y/100 (Count=6) is repeatedly selected by accuracy-, MCC-F1-, sensitivity-specificity-, and GPS1--2-based criteria, but it is not retained by GPS3, indicating weaker alignment with the balanced composite. Angle/Ring/Pauli-X/100 (Count=5) concentrates on accuracy/sensitivity-specificity and GPS1--2, while Star/Pauli-Z settings mainly enter through GPS3--4, reflecting an objective-dependent shift toward balanced criteria.
\begin{table*}[htpb]
\centering
\footnotesize
\caption{\textbf{Dataset 4:} Overlap of Top-5 configurations across metrics.}
\label{tab:d4_top5_overlap_values_c3}
\begin{adjustbox}{max width=\textwidth}
\begin{tabular}{l c c c c c c c c}
\hline
\textbf{Config (Enc / Arch / Meas / Shots)} & \textbf{Count} &
\textbf{Acc} & \textbf{MCC-F1} & \textbf{Sens-Spec} & \textbf{GPS1} & \textbf{GPS2} & \textbf{GPS3} & \textbf{GPS4} \\
\hline
QSample / Strong / Pauli-Y / 50 & 7 &
3(0.8000) & 2(0.7245) & 4(0.7737) & 3(0.8515) & 3(0.8160) & 1(0.7417) & 1(0.7554) \\
IQP / Strong / Pauli-Y / 100 & 6 &
2(0.8167) & 1(0.7444) & 2(0.7926) & 1(0.8665) & 1(0.8301) & -- & 2(0.7342) \\
Angle / Ring / Pauli-X / 100 & 5 &
1(0.8250) & 4(0.7127) & 1(0.8244) & 2(0.8526) & 2(0.8198) & -- & -- \\
Amplitude / Star / Pauli-Z / 50 & 4 &
-- & -- & 5(0.7088) & -- & 5(0.7033) & 2(0.7328) & 3(0.7232) \\
IQP / Star / Pauli-Z / 400 & 3 &
5(0.7167) & -- & -- & -- & -- & 3(0.7303) & 4(0.7211) \\
\hline
\end{tabular}
\end{adjustbox}
\end{table*}
\subsection{Discussion and takeaways}
HQNN selection is multi-criteria: encoding, architecture, and measurement jointly influence both peak performance and stability across the 625-run grid, whereas the shot budget in the tested range mainly acts as a secondary adjustment. This matters for healthcare tasks such as kidney disease prediction, where false positives and false negatives carry asymmetric costs and model choice should reflect error structure in addition to aggregate scores. Although the primary goal is design space exploration, the sweep also shows that multiple HQNN configurations achieve strong predictive performance, indicating that clinically relevant accuracy is attainable within compact hybrid settings when design factors are co-selected appropriately. Factor-averaged GPS summaries offer an efficient first-stage filter, but distributional views show that strong averages can coincide with high variance and heavier lower tails, motivating explicit robustness checks. Consistently, MCC--F1 and specificity-sensitivity diagnostics show that configurations with similar accuracy can exhibit different confusion-matrix asymmetries and operating-point tradeoffs that are not resolved by threshold tuning alone. Finally, the Top-5 overlap analysis (Count $\ge 3$) highlights configurations that remain competitive across objectives, reducing dependence on any single metric.

\noindent \textbf{Practical selection protocol:}
\begin{itemize}
    \item \textbf{Shortlist by intended objective:} Use GPS summaries to narrow candidates under the intended objective (accuracy-oriented vs balanced composites).
    \item \textbf{Stress-test robustness:} Verify stability using accuracy distributions across measurement choices and shot settings, prioritizing candidates with weaker lower tails.
    \item \textbf{Select using error structure:} Finalize using specificity-sensitivity and MCC--F1 diagnostics (with overlap recurrence), selecting operating regimes consistent with clinical priorities.
\end{itemize}

\section{Conclusion}\label{sec:conclusion}
We performed a systematic design space exploration of HQNNs for chronic kidney disease prediction, benchmarking 625 configurations under a unified protocol. The results show that interactions among encoding, topology, and measurement largely determine both performance and stability, while the tested shot range mainly provides a secondary adjustment that can reduce compute without materially changing accuracy when core design choices are well matched. For healthcare settings with asymmetric error costs, the study underscores that configuration selection should incorporate robustness and error structure, not only a single summary score. Overall, the proposed workflow combines composite-score screening with distributional and diagnostic validation, offering practical guidance for HQNN selection under deployment constraints.

\section*{Acknowledgment}
This work was supported in part by the NYUAD Center for Quantum and Topological Systems (CQTS), funded by Tamkeen under the NYUAD Research Institute grant CG008.

\begin{spacing}{0.92}
\bibliographystyle{IEEEtran}
\bibliography{refs}
\end{spacing}

\end{document}